\documentclass[conference]{IEEEtran}
\usepackage{cite}
\usepackage{amsmath,amssymb,amsfonts}
\usepackage{graphicx}
\usepackage{textcomp}
\usepackage{xcolor}
\usepackage{booktabs}
\usepackage{multirow}
\usepackage{fancyhdr}
\usepackage{url}
\usepackage{gensymb}

\graphicspath{{figures/}{../../data/processed/}}
\def\BibTeX{{\rm B\kern-.05em{\sc i\kern-.025em b}\kern-.08em
    T\kern-.1667em\lower.7ex\hbox{E}\kern-.125emX}}

\title{RL-Driven Sustainable Land-Use Allocation\\for the Lake Malawi Basin}

\author{\IEEEauthorblockN{Ying Yao}
\IEEEauthorblockA{\textit{Georgia Institute of Technology}\\
yyao374@gatech.edu}}

\begin{document}
\maketitle
\pagestyle{plain}
\thispagestyle{fancy}

\begin{abstract}
Unsustainable land-use practices in ecologically sensitive regions
threaten biodiversity, water resources, and the livelihoods of millions.
We present a deep reinforcement learning (RL) framework that optimises
land-use allocation in the Lake Malawi Basin to maximise total ecosystem
service value (ESV). Biome-specific ESV coefficients are derived via
benefit transfer from Costanza~et~al.\ and locally anchored to a Malawi
wetland valuation, with a regenerative-agriculture uplift on the crops
coefficient as the headline policy setting. A Proximal Policy
Optimization (PPO) agent operates over a $50\!\times\!50$ cell grid at
500\,m resolution, transferring land-use pixels under a joint,
cell-level action mask. The reward combines per-cell ecological value
with spatial-coherence terms---contiguity bonuses for forest, cropland,
and built patches; a buffer penalty paired with a structural
riparian mask that forbids new high-impact placements next to water;
and a riparian-tree bonus incentivising forest cover along the
shoreline. The trained policy raises total ESV, consolidates forest
and urban cells into contiguous blocks, and actively restores
high-impact cells away from water, demonstrating utility as a
scenario-analysis tool for environmental planning.
\end{abstract}


\section{Introduction}

Lake Malawi, the third-largest lake in Africa, harbors extraordinary
aquatic biodiversity---including more than 1{,}000 endemic cichlid fish
species---and sustains the livelihoods of over five million people through
fisheries, agriculture, and freshwater supply. Yet the basin faces
accelerating pressures from deforestation, agricultural expansion, and
unplanned urbanization, all of which degrade the ecosystem services on
which local communities depend~\cite{millennium_ecosystem_2005}. Making
informed land-use decisions in such contexts requires a quantitative
framework that can (i)~value the diverse services provided by different
biome types, (ii)~account for spatial interactions among adjacent land
parcels, and (iii)~explore the consequences of alternative development
strategies.

\subsection{Ecosystem Service Valuation}
Ecosystem service value (ESV) provides a monetary accounting of the
benefits that ecosystems deliver to human well-being---provisioning
(food, freshwater), regulating (climate, water purification), and
cultural (recreation, aesthetic) services.
Costanza~et~al.~\cite{costanza_value_1997} pioneered the global
estimate at US\$33~trillion/yr (1995~dollars), later revised to
US\$125~trillion/yr (2007~dollars) using the expanded
ESVD~\cite{de_groot_global_2012,costanza_changes_2014}. The benefit
transfer method assigns unit values (USD/ha/yr) per biome type,
making land-use trade-offs explicit, e.g.\ wetland-to-cropland
conversion raises provisioning output while sacrificing regulating
services such as flood control and nutrient
filtration~\cite{costanza_changes_2014}. This framework grounds our
reward function.

\subsection{Deep Reinforcement Learning for Land-Use Optimization}
Reinforcement learning (RL) formulates sequential decision-making as an
agent interacting with an environment to maximize cumulative
reward~\cite{sutton_reinforcement_2018}. Recent advances in deep RL (DRL)
have demonstrated the potential of this paradigm for spatial planning
tasks. Zheng~et~al.~\cite{zheng_spatial_2023} applied DRL with graph
neural networks to urban community planning, optimizing land-use and road
layouts to maximize the ``15-minute city'' accessibility metric.
Shen~et~al.~\cite{shen_urban_2024,shen_optimizing_2025} proposed a
DRL-based carbon emission mitigation strategy integrating Points of
Interest and transportation data, later extending it to Hangzhou with
PPO over cell-level land-use proportions and achieving up to 15\%
carbon-emission reductions over baseline configurations.

These studies share a common formulation---a gridded study area,
cell-level land-use fractions as the RL state, and actions that
incrementally transfer between classes. We adopt this template but
shift the target from urban carbon-emission reduction to ESV
maximisation in an ecological basin, with different dominant classes
(forest, wetland, rangeland) and conservation priorities (habitat
connectivity, riparian buffers).

\subsection{Contributions}
This paper contributes:
\begin{enumerate}
    \item An RL framework for ESV-driven land-use optimisation in a
          non-urban, ecologically critical setting, the Lake Malawi
          Basin.
    \item A multi-objective reward combining benefit-transfer ESV
          coefficients, contiguity bonuses for habitat
          connectivity~\cite{fahrig_effects_2003}, a buffer penalty on
          high-impact land use near water, and a riparian-tree bonus
          rewarding restorative forest placement along the
          shoreline~\cite{lowrance_riparian_1984}.
    \item A joint, cell-level action mask that validates each
          (cell, source, target) tuple in one shot---source present,
          target with room, classes distinct---so every sampled action
          is immediately executable; the mask also encodes a
          structural riparian-buffer rule forbidding new high-impact
          targets at water-adjacent cells.
    \item Evidence that the trained policy internalises all reward
          terms---raising contiguity, leaving a persistent restorative
          bias on the buffer penalty, and concentrating trees along
          the shoreline---complemented by two reward-design ablations.
\end{enumerate}

\section{Materials and Methods}

\subsection{Study Region}
\begin{figure}[htbp]
\centerline{\includegraphics[width=0.42\textwidth]{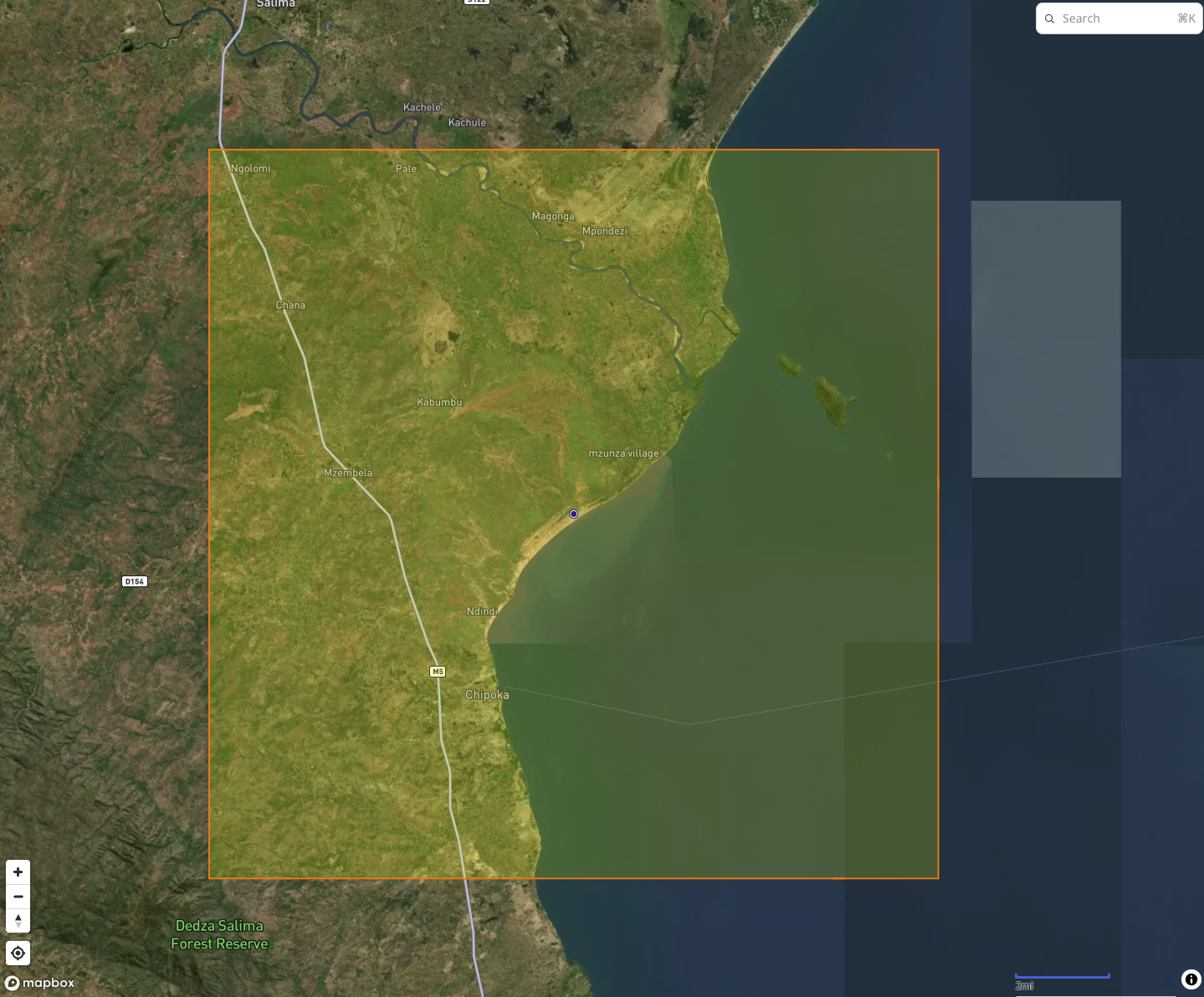}}
\caption{Satellite view of the $25\times25$\,km study region on the
         western shore of Lake Malawi. The green overlay denotes the area
         of interest.}
\label{fig:study_region}
\end{figure}

Lake Malawi is located between Malawi, Mozambique, and Tanzania and is the
ninth-largest lake in the world by area. The study focuses on a
$25\,\text{km}\times25\,\text{km}$ area on the western shore of the lake,
centered at latitude $-13.934564\degree$, longitude $34.542859\degree$
(Fig.~\ref{fig:study_region}). This region was selected for its diverse
land-cover composition: water (45.2\%), trees (0.3\%), flooded/wetlands
(2.4\%), crops (24.5\%), built area (7.9\%), bare ground (0.1\%), and rangeland
(19.7\%)---providing a rich testbed for multi-class land-use optimization.

\subsection{Data Sources and Preprocessing}
\subsubsection{Land Cover}
Land-cover data were obtained from the ESA WorldCover
product~\cite{zanaga_esa_2022}, derived from Sentinel-2 imagery at 10\,m
resolution for the year 2024. Nine land-cover classes are present in the
study area (Table~\ref{tab:esv}). To reduce dimensionality, we
downsample to 100\,m resolution by taking the modal class within each
$10\!\times\!10$ pixel block (10\% sampling). The downsampled map is then
organized into a grid of $50\!\times\!50$ cells, each cell comprising
$5\!\times\!5$ pixels at 100\,m, yielding a 500\,m effective resolution
per cell. Each cell stores a 9-dimensional vector of pixel counts per
land-cover class.

\subsubsection{Evapotranspiration}
Evapotranspiration (ET) data are drawn from the MODIS MOD16A3GF
product~\cite{running_mod16_2021} at 500\,m resolution for 2024, accessed
via the Planetary Computer API. Yearly ET values (kg/m$^2$/yr)
are computed per land-cover class by area-weighted averaging over the
study region (Table~\ref{tab:et}). Trees exhibit the highest ET rate,
consistent with their canopy transpiration, while bare ground and water
surfaces sit at the lower end. While ET is
not directly used in the reward function, it serves as an environmental
constraint: episodes terminate early if the cumulative ET decrease exceeds
a configurable tolerance, preventing allocations that would severely
compromise water cycling.

\begin{table}[htbp]
\caption{Per-Class Evapotranspiration (2024, MOD16A3GF)}
\label{tab:et}
\centering
\begin{tabular}{@{}lcc@{}}
\toprule
Class & Type & ET (kg/m$^2$/yr) \\
\midrule
Water         & Protected  & 616.93 \\
Trees         & Modifiable & 933.57 \\
Flooded       & Protected  & 767.62 \\
Crops         & Modifiable & 675.66 \\
Built Area    & Modifiable & 648.95 \\
Bare Ground   & Modifiable & 591.53 \\
Snow/Ice      & Protected  & ---    \\
Clouds        & Protected  & 845.87 \\
Rangeland     & Modifiable & 745.94 \\
\bottomrule
\end{tabular}
\end{table}

\subsection{Ecosystem Service Valuation}\label{sec:esv}
We adopt a benefit transfer approach to assign ESV coefficients to each
land-cover class. Inter-biome ratios are derived from the updated unit
values in Costanza~et~al.~\cite{costanza_changes_2014}, which synthesizes
over 300 case studies in the ESVD~\cite{de_groot_global_2012}. To anchor
these global ratios to the local context, we calibrate against a primary
valuation study for the Lake Chiuta wetland in southern Malawi, which
estimated the gross financial value of inland wetland services at
US\$554/ha/yr~\cite{zuze_economic_2013}. Ratios for each biome type are
then scaled relative to this anchor (Table~\ref{tab:esv}).

Four classes---water, flooded/wetlands, snow/ice, and clouds---are
designated as \emph{protected}: the RL agent can neither observe nor modify
them, reflecting physical or regulatory constraints. The remaining five
\emph{modifiable} classes (trees, crops, built area, bare ground,
rangeland) constitute the agent's observation and action space.

Our headline configuration applies a $1.35\times$ regenerative-agriculture
uplift to the crops coefficient based on the estimation from new water paradigm~\cite{mueller_global_nodate}
; the base value is used only in the Appendix~A ablation. All ESV coefficients are
min-max normalized to $[0,1]$ before use in the reward function,
eventually among the modifiable classes, the
normalized values rank as: regenerative crops (0.29) $>$ built area (0.26) $>$ trees
(0.21) $>$ rangeland (0.16) $>$ bare ground (0.00).

\begin{table}[htbp]
\caption{Land-Cover Classes and Ecosystem Service Values. Crops
carries the $1.35\times$ regenerative-agriculture uplift used in the
headline reward; the base value $246$ is used only in the Appendix~A
ablation.}
\label{tab:esv}
\centering
\begin{tabular}{@{}llcc@{}}
\toprule
Class & Type & ESV & Ratio \\
      &      & (USD/ha/yr) & \\
\midrule
Water         & Protected  & 554  & 1.00 \\
Trees         & Modifiable & 238  & 0.43 \\
Flooded       & Protected  & 1{,}136 & 2.05 \\
Crops         & Modifiable & 246 * 1.35 = 332  & 0.60 \\
Built Area    & Modifiable & 295  & 0.53 \\
Bare Ground   & Modifiable & 0    & 0.00 \\
Snow/Ice      & Protected  & 0    & ---  \\
Clouds        & Protected  & 0    & ---  \\
Rangeland     & Modifiable & 184  & 0.33 \\
\bottomrule
\end{tabular}
\end{table}

\subsection{Training Dataset Preparation}
The $50\!\times\!50$ cell grid is partitioned into non-overlapping
$10\!\times\!10$ sub-patches, yielding 25 samples. These are split 70/30
into training and test sets, with indices
fixed by a random seed. To mitigate overfitting on the small sample count,
we apply spatial data augmentation: each original patch is randomly shifted
by $\pm 2$ cells in both row and column directions for $n_{\text{aug}}=5$
rounds, expanding the effective training set by a factor of 6. Patches
with a modifiable land fraction below 10\% are rejected during episode
initialization to ensure the agent has meaningful action opportunities.

\subsection{Reinforcement Learning Framework}

\begin{figure*}[htbp]
\centering
\includegraphics[width=0.85\textwidth]{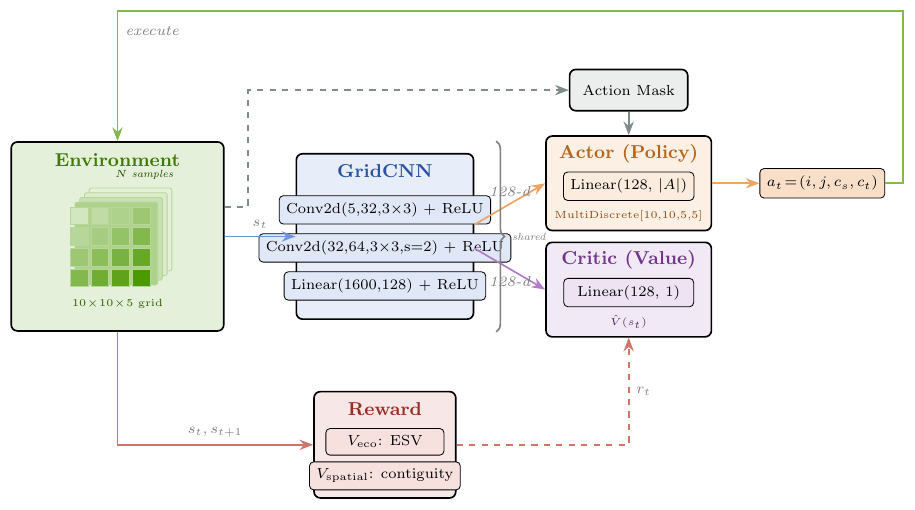}
\caption{Overview of the proposed RL framework. The agent observes a
         $10\!\times\!10\!\times\!5$ land-use fraction grid, extracts
         spatial features via a shared GridCNN, and produces both a
         masked policy (Actor) and state value estimate (Critic). The
         environment executes the selected action and returns a reward
         combining ESV change and spatial coherence metrics.}
\label{fig:architecture}
\end{figure*}

\subsubsection{State Space}
The observation at each timestep is a three-dimensional tensor
$\mathbf{s}\in[0,1]^{M\times M\times K}$, where $M=10$ is the grid
dimension and $K=5$ is the number of modifiable land-cover classes. Each
element $s_{i,j,k}$ represents the fraction of cell $(i,j)$ occupied by
modifiable class~$k$, computed as the pixel count divided by the total
pixels per cell ($N_p = 25$). Protected classes are excluded from the
observation entirely (i.e., the agent has no channel for water or wetlands),
however, the protected water fraction is retained internally for computing
the buffer zone penalty.

\subsubsection{Action Space and Masking}
Each action selects a single cell and a transfer direction, and is encoded
as a joint \texttt{Discrete} index $a \in [0, M^2 K^2)$ that decodes into a
4-tuple $(i, j, c_{\text{src}}, c_{\text{tgt}})$ via
$(i,j,c_{\text{src}},c_{\text{tgt}}) = \mathrm{unravel}(a, (M,M,K,K))$.
An action transfers $\Delta = 5$ pixels from source class $c_{\text{src}}$
to target class $c_{\text{tgt}}$ in cell $(i,j)$; the transfer is clamped
so that the source fraction cannot fall below 0 and the target fraction
cannot exceed 1, guaranteeing zero-sum conservation within each cell by
construction.

To keep the policy exclusively on feasible actions, we employ action
masking via MaskablePPO~\cite{huang_closer_2022,raffin_stable_2021}. At
each step, a joint boolean mask of shape $(M, M, K, K)$ is assembled and
flattened into a length-$M^2 K^2$ vector. A tuple $(i,j,c_{\text{src}},
c_{\text{tgt}})$ is valid iff both conditions hold:
\begin{itemize}
    \item \textbf{Per-cell feasibility}:
          $s_{i,j,c_{\text{src}}} > 0$,
          $s_{i,j,c_{\text{tgt}}} < 1$, and
          $c_{\text{src}} \neq c_{\text{tgt}}$ --- i.e., the source
          class is present in that specific cell, the target class has
          room there, and the two classes differ.
    \item \textbf{Riparian protection}: if cell $(i,j)$ is adjacent to a
          protected water cell, then
          $c_{\text{tgt}} \notin \{\text{crops},\,\text{built area}\}$
          --- the two highest-impact classes for water
          quality~\cite{lowrance_riparian_1984}. The agent therefore
          cannot \emph{expand} either class toward water, but
          restoration moves \emph{out of} water-adjacent cells (e.g.\
          built~$\rightarrow$~trees) remain legal.
\end{itemize}

\subsubsection{Reward Function}\label{sec:reward}
The reward at step $t$ is the change in total value:
\begin{equation}
    r_t = V(\mathbf{s}_{t+1}) - V(\mathbf{s}_t)
\end{equation}
where the total value $V$ combines ecological value with spatial
shaping:
\begin{equation}\label{eq:total_value}
    V(\mathbf{s}) = V_{\text{eco}}(\mathbf{s})
                  + V_{\text{spatial}}(\mathbf{s})
\end{equation}

The \textbf{ecological value} is the sum of ESV-weighted fractions across
all cells:
\begin{equation}\label{eq:eco}
    V_{\text{eco}}(\mathbf{s})
      = \sum_{i=1}^{M}\sum_{j=1}^{M}\sum_{k=1}^{K}
        s_{i,j,k}\;\tilde{e}_k
\end{equation}
where $\tilde{e}_k$ is the min-max normalized ESV coefficient for
modifiable class~$k$.

The \textbf{spatial value} captures landscape-level ecological
coherence:
\begin{equation}\label{eq:spatial}
    V_{\text{spatial}}(\mathbf{s})
      = w_T\,C_T + w_C\,C_C + w_B\,C_B - w_W\,P_W + w_R\,B_R
\end{equation}
where $C_T$, $C_C$, $C_B$ are contiguity scores for trees, crops, and
built area respectively, $P_W$ is the water buffer penalty, and $B_R$ is
a riparian-tree bonus that rewards forest cover adjacent to water. Each
contiguity score measures how much a given class is spatially clustered:
\begin{equation}\label{eq:contiguity}
    C_c = \ln\!\Bigl(1 + \sum_{i,j} s_{i,j,c}\;
          (\mathbf{K} \ast \mathbf{s}_{:,:,c})_{i,j}\Bigr)
\end{equation}
Here $\mathbf{K}$ is the 4-connected neighbor kernel
$\bigl[\begin{smallmatrix}0&1&0\\1&0&1\\0&1&0\end{smallmatrix}\bigr]$
and $\ast$ denotes 2D convolution with zero-padded boundaries. The
$\ln(1+\cdot)$ transformation compresses the range to prevent large
contiguous patches from dominating the reward signal.

The \textbf{buffer zone penalty} discourages high-impact land use
(crops and built area) adjacent to water:
\begin{equation}\label{eq:buffer}
    P_W = \ln\!\Bigl(1 + \sum_{i,j}
          (s_{i,j,\text{crop}} + s_{i,j,\text{built}})\;
          (\mathbf{K} \ast \mathbf{w})_{i,j}\Bigr)
\end{equation}
where $\mathbf{w}\in[0,1]^{M\times M}$ is the protected water fraction
map, encoding the principle that riparian buffers reduce runoff and
protect aquatic ecosystems~\cite{lowrance_riparian_1984}. The
\textbf{riparian-tree bonus} mirrors $P_W$ with opposite sign,
rewarding forest cover at water-adjacent cells:
\begin{equation}\label{eq:riparian}
    B_R = \ln\!\Bigl(1 + \sum_{i,j}
          s_{i,j,\text{tree}}\;
          (\mathbf{K} \ast \mathbf{w})_{i,j}\Bigr)
\end{equation}
$B_R$ complements the structural riparian mask by providing a
positive signal for restorative tree placement.
Headline-configuration weights are listed in
Table~\ref{tab:spatial_weights}; $w_W$ is annealed from $1$ to $6$
over the first $60\%$ of training so the policy first learns
profitable transfers under a mild buffer penalty.

\begin{table}[htbp]
\caption{Spatial Reward Weights. $w_W$ is annealed linearly from $1$
to $6$ over the first $60\%$ of training; the other weights are held
fixed.}
\label{tab:spatial_weights}
\centering
\begin{tabular}{@{}lcc@{}}
\toprule
Component & Symbol & Weight \\
\midrule
Tree contiguity bonus       & $w_T$ & 1.0 \\
Crop contiguity bonus       & $w_C$ & 4.0 \\
Built-area contiguity bonus & $w_B$ & 2.0 \\
Water buffer penalty        & $w_W$ & 6.0 (anneal $1{\rightarrow}6$) \\
Riparian-tree bonus         & $w_R$ & 5.0 \\
\bottomrule
\end{tabular}
\end{table}

\subsubsection{Termination Conditions}
An episode terminates when any of the following conditions is met:
\begin{enumerate}
    \item \textbf{Step limit}: the agent has taken $T_{\max}=500$ steps.
    \item \textbf{ET constraint}: the fractional decrease in total
          evapotranspiration from the initial state exceeds a tolerance
          $\tau_{\text{ET}}=1.0$ (i.e., ET may decrease by up to 100\%).
    \item \textbf{Stagnation}: the agent has produced
          $n_{\text{noop}}=10$ consecutive no-op actions, indicating it
          can find no further improving transfers.
    \item \textbf{Saturation}: the joint mask admits no valid action in
          the current state (every cell is either empty or saturated),
          signalling that the environment is terminal by construction.
\end{enumerate}

\subsection{Network Architecture}
We employ a custom CNN feature extractor (\textsc{GridCNN}) that replaces
the default MLP feature extractor in Stable-Baselines3. The architecture
processes the $K\!\times\!M\!\times\!M$ observation (channels first) as
follows:
\begin{enumerate}
    \item $\text{Conv2d}(K{=}5,\;32,\;3{\times}3,\;\text{stride}=1,\;\text{pad}=1) \rightarrow \text{ReLU}$
    \item $\text{Conv2d}(32,\;64,\;3{\times}3,\;\text{stride}=2,\;\text{pad}=1) \rightarrow \text{ReLU}$
    \item $\text{Flatten} \rightarrow \text{Linear}(1600,\;128) \rightarrow \text{ReLU}$
\end{enumerate}
The first convolutional layer preserves the $10\!\times\!10$ spatial
dimensions with a $3\!\times\!3$ receptive field structurally aligned
with the 4-connected contiguity kernel $\mathbf{K}$, providing an
inductive bias for the local-neighbour relationships that drive the
spatial reward. The second layer (stride~2) reduces to
$5\!\times\!5$ and captures multi-cell cluster patterns. The
resulting 128-dim feature vector is shared between the actor and
critic heads.

\subsection{Training Configuration}
The agent is trained using MaskablePPO~\cite{schulman_proximal_2017,
huang_closer_2022} from the SB3-Contrib library~\cite{raffin_stable_2021}
for $1{,}500{,}000$ timesteps, with metrics logged to
Weights~\&~Biases. A low entropy coefficient ($0.005$, annealed to
$0.001$ over the last $33\%$), a small learning rate
($5\!\times\!10^{-5}$, decayed to $5\!\times\!10^{-6}$ over the same
tail), and a tight maximum gradient norm ($0.25$) encourage the policy
to commit while keeping updates firmly inside the PPO trust region. An
episode-initialisation filter rejects training samples whose initial
total value \begin{math}V_0\end{math} falls below $1.0$ (typical of grids dominated by protected
water/wetland cells), ensuring the agent always faces non-trivial
states. The full hyperparameter table and a note on numerical handling
of the 2500-dim softmax are deferred to Appendix~B.

\section{Results}
We report the behaviour of the trained policy under the headline reward
configuration (Section~\ref{sec:reward}) and the ESV coefficients of
Table~\ref{tab:esv}. Two reward-design ablations---\emph{eco-only}
($V_{\text{spatial}}$ zeroed) and \emph{spatial-without-regen} (base
crop ESV)---are reported in Appendix~A for context.

\subsection{Land-Use Allocation}\label{sec:primary_results}
\begin{figure}[htbp]
\centerline{\includegraphics[width=0.48\textwidth]{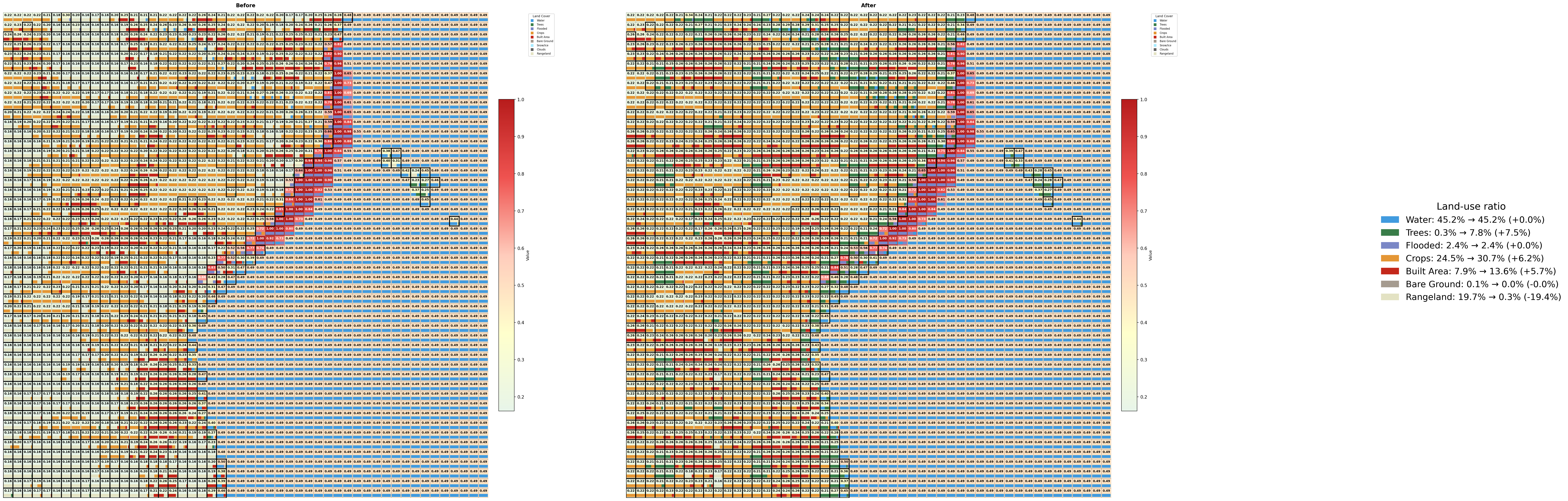}}
\caption{Initial (left) and post-optimisation (right) land-use
         allocation under the headline reward configuration.}
\label{fig:primary}
\end{figure}

After training, the agent is evaluated on both non-augmented (original) training and test
patches; the resulting cell-level allocations are reconstructed onto
the full $50\!\times\!50$ grid and visualised as side-by-side
before/after heatmaps (Fig.~\ref{fig:primary}). Each cell displays
horizontal bars for land-use type fractions, and the background colour
encodes the cell's total ecosystem value (red = high, white = low).
Black-bordered cells indicate locations modified by the agent. The
right-hand legend reports the global before/after ratio of each
land-use type with its percentage-point change, summarising
the aggregate shift in the allocation.

Three behaviours dominate the ``after'' map. First, the agent expands
cropland at the expense of bare ground and rangeland---consistent
with the regenerative-agriculture uplift that makes crops the
highest-value modifiable class (0.29)---and consolidates them via
the $w_C{=}4$ bonus. Second, built-area and tree growth is clustered
rather than sprawling, reflecting the contiguity bonuses. Third,
the shoreline shows visible built/crop~$\rightarrow$~trees transfers:
the riparian mask forbids \emph{new} high-impact placements while the
buffer penalty and riparian-tree bonus jointly drive \emph{removal}
of pre-existing violations. The result is an allocation that
simultaneously raises per-cell ESV, respects contiguity, and cleans
up the riparian buffer. A region-by-region comparison across
ablations is provided in Appendix~C.

\subsection{Training Diagnostics}\label{sec:metric_analysis}
We examine the training curves logged to Weights~\&~Biases to verify
that the learned behaviour reflects genuine convergence rather than an
artifact of reward scaling or update-step instability.

\subsubsection{Return and episode length}
\begin{figure}[htbp]
\centerline{\includegraphics[width=0.48\textwidth]{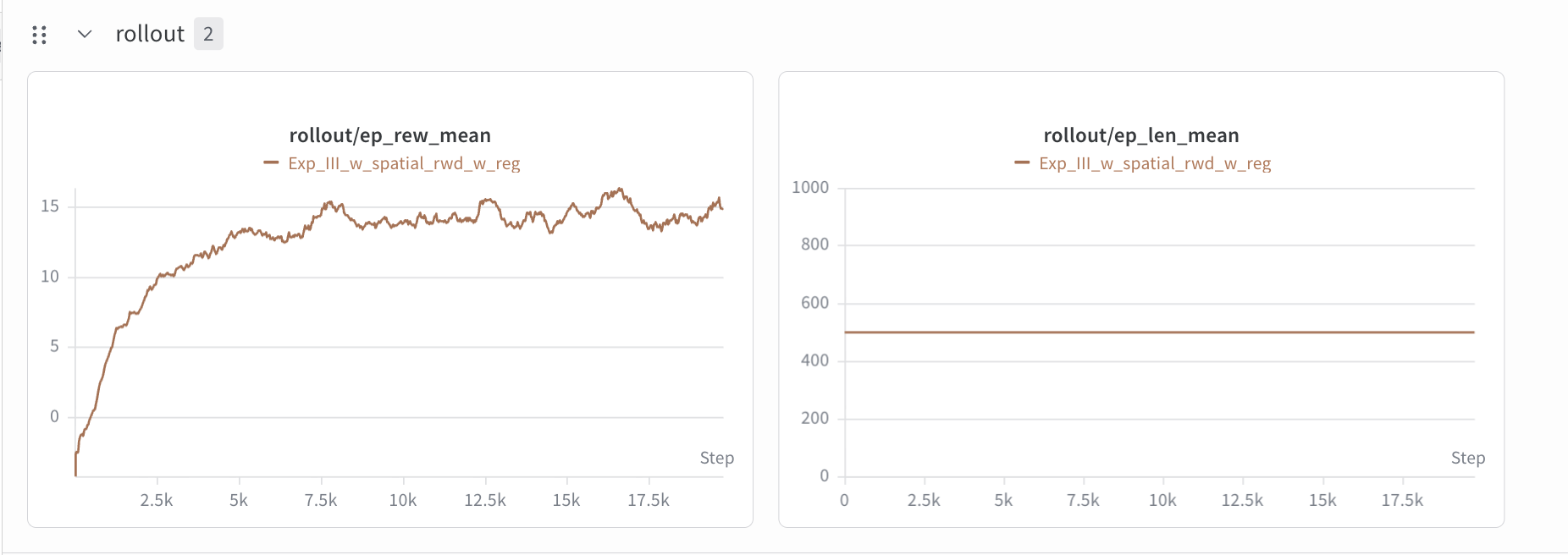}}
\caption{Training return (left) and episode length (right).}
\label{fig:metric_rwd}
\end{figure}

The mean episode return (\texttt{rollout/ep\_rew\_mean}) rises from a
slightly negative initial value, climbs sharply through the first
$\sim$5k rollout steps, and plateaus in the 13--16 band for the
remainder of training (Fig.~\ref{fig:metric_rwd}, left). Mean episode
length (Fig.~\ref{fig:metric_rwd}, right) pins at the $T_{\max}=500$
cap throughout training, indicating that the joint action mask
consistently admits legal improving transfers and the agent uses its
full step budget rather than terminating early on no-op cascades.
Detailed critic and loss-curve diagnostics---including the explained
variance, entropy-coefficient anneal effect on the policy entropy, and
the learning-rate decay effect on the PPO clip fraction---are deferred
to Appendix~B.

\subsubsection{Spatial reward components}
\begin{figure}[!t]
\centerline{\includegraphics[width=0.48\textwidth]{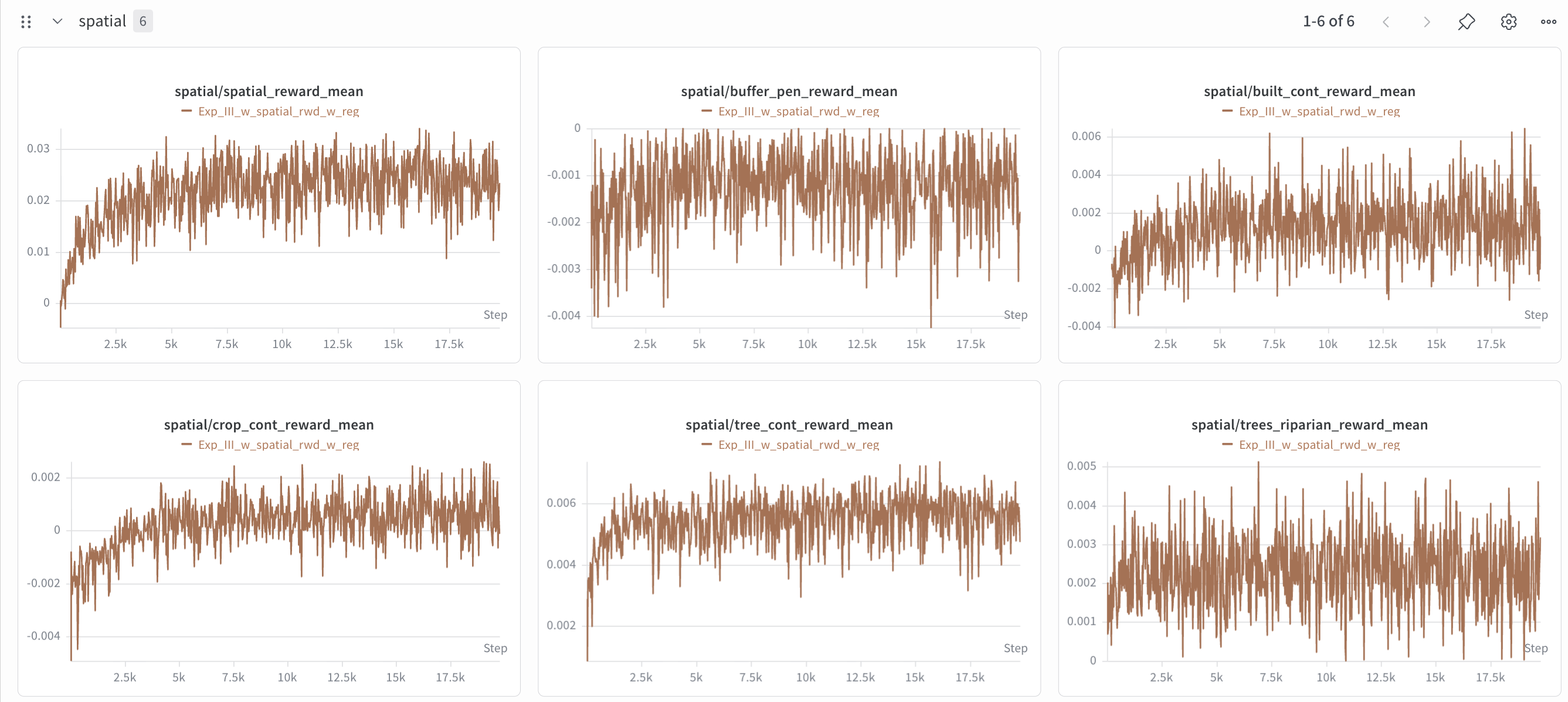}}
\caption{Spatial reward components during training.}
\label{fig:metric_spatial}
\end{figure}

Fig.~\ref{fig:metric_spatial} decomposes the spatial reward into the
five channels of Eq.~\ref{eq:spatial}. Tree and built-area
contiguity rewards rise steadily, confirming that the agent gradually
consolidates forest and urban cells. Crop contiguity drifts from
slightly negative toward zero with a small positive bias---crops
start from a sparse baseline. The buffer-penalty reward stays near
zero with a persistent slight-negative bias: the riparian mask
forbids \emph{new} high-impact placements at water-adjacent cells,
so even as $w_W$ ramps from $1$ to $6$ the penalty only registers a
small restorative gradient as the agent removes pre-existing
violations (built/crop~$\rightarrow$~trees). The riparian-tree bonus
($w_R{=}5$) hovers in a small positive range, supplying the
complementary restorative incentive. Active spatial shaping is
therefore carried by the tree- and built-contiguity bonuses together
with the riparian-tree bonus.

 \subsection{Baseline Comparison}\label{sec:baselines}
To situate the trained policy against non-learning references, we
compare two baselines with the above experiment on the test
split: a uniform \emph{Random} policy that samples each step from the
valid action set, and a \emph{Greedy} oracle that picks
$\arg\max_a \Delta V$ via simulated 1-step rollouts of \begin{math}r_t\end{math}. 
Note that these 3 methods share the same action space, mask and value estimation.
Of the 48 augmented test samples, 24 are \emph{trivial}---fully covered by
protected classes with $V_0\!\le\!1$ and therefore structurally
un-improvable; we filter them out and report results on the 24
\emph{effective} samples only.

\begin{figure}[!t]
\centerline{\includegraphics[width=0.48\textwidth]{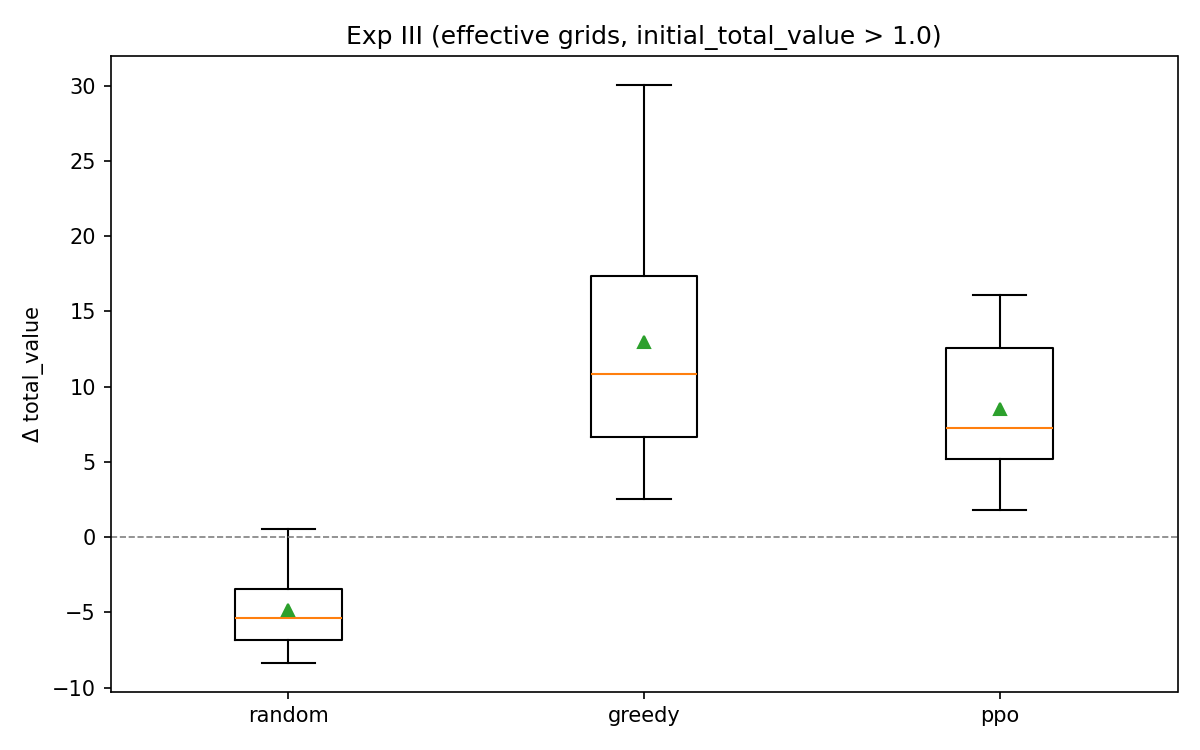}}
\caption{$\Delta V$ distribution per method on the 24 effective test
grids ($V_0\!>\!1$). Triangles mark means.}
\label{fig:baseline_box}
\end{figure}

Across the 24 effective grids (Fig.~\ref{fig:baseline_box}), PPO
improves total ESV on every episode (mean $\Delta V = 8.50 \pm 4.56$,
success rate $1.0$) whereas Random consistently \emph{destroys} value
(mean $\Delta V = -4.82 \pm 2.49$, success rate $0.04$). PPO
therefore clears the no-intelligence floor decisively---an
end-to-end policy that has internalised the eco-plus-spatial reward
structure. Against the Greedy oracle, however, PPO closes only
$\sim\!66\,\%$ of the available gain ($8.50$ vs.\ $12.95 \pm 8.76$).
Both methods reach essentially the same final eco mass
($27.59$ vs.\ $27.66$); the residual gap sits in the spatial term,
where PPO incurs much more buffer-penalty mass
($0.357$ vs.\ $0.000$) and slightly weaker crop and built contiguity
($5.23$ vs.\ $5.45$ and $3.89$ vs.\ $4.20$), only partly compensated
by higher tree contiguity ($2.30$ vs.\ $2.08$) and offset further by
lower riparian-tree mass ($0.29$ vs.\ $0.49$).

\begin{figure}[!t]
\centerline{\includegraphics[width=0.48\textwidth]{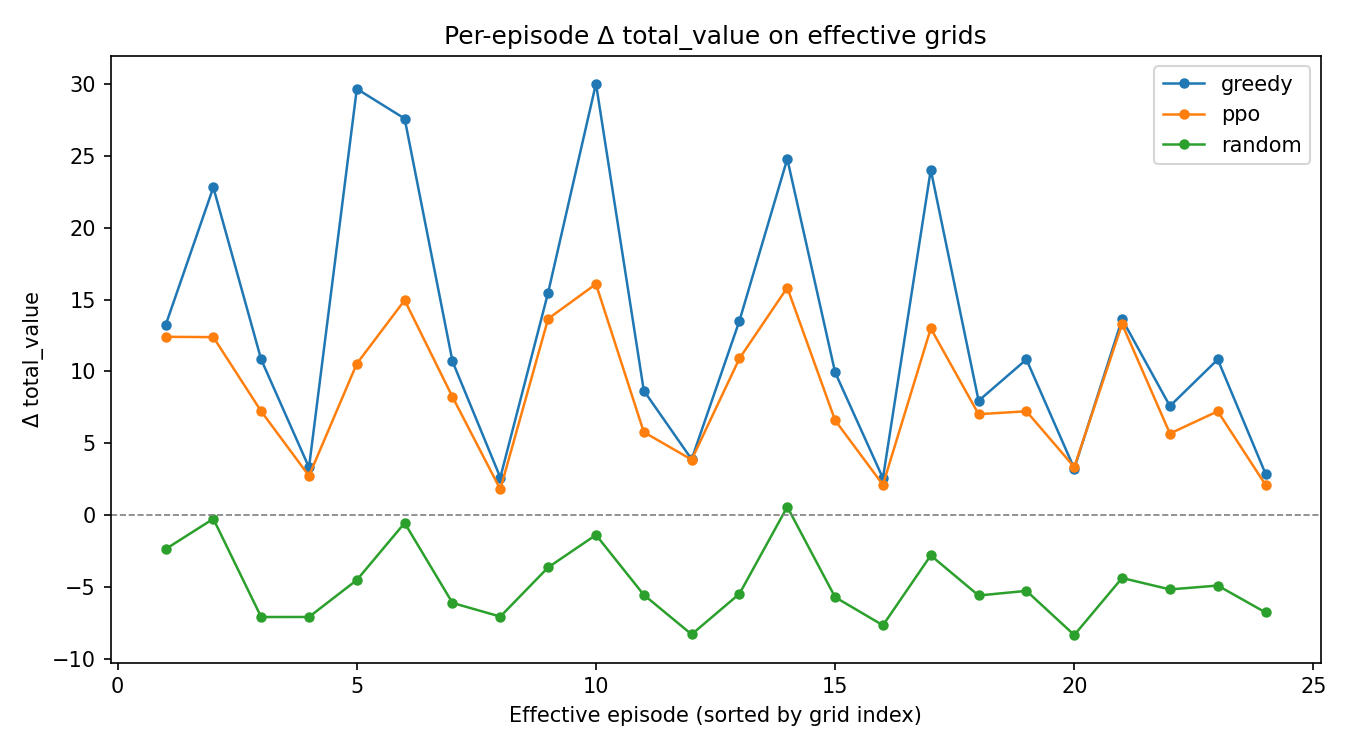}}
\caption{Per-episode $\Delta V$ on the 24 effective grids, ordered by
grid index. Greedy dominates PPO grid-by-grid (24/24); both dwarf
Random throughout.}
\label{fig:baseline_line}
\end{figure}

The per-episode trace in Fig.~\ref{fig:baseline_line} shows the
ordering is stable rather than mean-driven---Greedy~$>$~PPO~$\gg$~Random
holds on all 24 effective grids---and that PPO tracks Greedy's per-grid
shape closely, suggesting the two policies converge to qualitatively
related local optima of the same reward but at different operating
points on the eco--spatial trade-off.

\textbf{Why does PPO underperform Greedy?}\quad
We identify three plausible directions worth investigating.
First, \textbf{partial observability}: PPO's large
buffer-penalty gap suggests that, with only the five modifiable
classes exposed in the observation, the policy probably cannot directly
perceive water adjacency---the protected-class indicator and its
neighbourhood convolution $\mathit{water\_nb}$ drive the buffer
term but are withheld from the agent. Exposing $\mathit{water\_nb}$
as an additional observation channel, or injecting it into the
GridCNN's intermediate feature map, would possibly let the policy condition
on the same geometry that shapes the reward.
Second, \textbf{residual non-convergence}: although the value
function attains near-perfect explained variance,
$\lvert\text{pg\_loss}\rvert\!\approx\!5\!\times\!10^{-3}$ and policy
entropy still retain non-trivial mass at the end of training,
indicating that a longer training horizon with sharper entropy and
learning-rate decay is likely to further close part of the gap.
Third, \textbf{exploration deficit on a $2{,}500$-way action head}:
a behaviour-cloning warm-start from Greedy rollouts followed by PPO
fine-tuning would serve as a discriminating diagnostic---if the
BC-initialised policy matches or exceeds Greedy, the bottleneck lies
in RL training; if not, the enhancement should be more focused on either the
observation or the reward design.

\section{Discussion and Conclusion}

\subsection{Limitations and Discussion}

\textbf{Reward design as policy encoding.}\quad
The contiguity bonuses, buffer penalty, and riparian-tree bonus
translate ecological principles---habitat
connectivity~\cite{fahrig_effects_2003} and riparian
protection~\cite{lowrance_riparian_1984}---into differentiable
objectives the agent can optimise through gradient-based learning. This
is more flexible than hard constraints alone because the weights can
be tuned to reflect varying policy priorities without modifying the
environment logic; the Appendix~A ablations (Eco-only vs.\ Spatial-only
vs.\ Spatial+Regen) illustrate how per-term weights and ESV coefficients
together steer the policy toward different planning outcomes.

Several limitations should be acknowledged:

\begin{itemize}
    \item \textbf{PPO underperforms a one-step Greedy oracle}: on
          the headline test grids PPO closes only $\sim\!66\,\%$ of
          Greedy's ESV gain ($\Delta V\!=\!8.50$ vs.\ $12.95$) and
          retains $\text{buf\_pen}\!=\!0.357$ where Greedy reaches
          exact zero. We attribute this to partial observability of
          the protected-class layer, residual training-time
          non-convergence, and exploration deficit on the
          $2{,}500$-way action head, and outline a discriminating
          behaviour-cloning experiment in Sec.~\ref{sec:baselines}.
    \item \textbf{Single study region}: results are demonstrated on one
          geographic area; generalisation to other regions with
          different land-cover distributions remains to be validated.
    \item \textbf{Reward-component balance under ESV shifts}: the
          spatial weights ($w_T, w_C, w_B, w_W, w_R$) and the $w_W$
          curriculum are held fixed under the regenerative-agriculture
          crop-ESV uplift; we do not re-scale $w_C$ to compensate, and
          whether auto-calibration is needed when ESV coefficients
          change remains open.
\end{itemize}

\subsection{Future Work}
Several directions are worth pursuing:
(i)~conducting formal ablation studies on the GridCNN versus MLP
feature extractor and on individual spatial reward components;
(ii)~benchmarking against established optimization baselines;
and (iii)~generalizing the framework to other ecologically sensitive regions
beyond Lake Malawi and enriching the reward design with other essential ecological principles 
to validate its effectiveness across diverse land-cover distributions and policy contexts.

\subsection{Conclusion}
We presented an RL framework for ESV-driven land-use allocation in
the Lake Malawi Basin. Combining benefit-transfer ESV coefficients
with spatial-coherence rewards---contiguity bonuses, a structural
riparian action mask, a buffer penalty, and a riparian-tree
bonus---the framework produces ecologically informed plans that
balance per-cell value with landscape-level objectives. The trained
policy reliably raises total ESV, consolidates forest and urban
cells into contiguous blocks, and restores high-impact cells away
from water; Appendix~A ablations confirm that the framework responds
meaningfully to changes in spatial shaping and ESV coefficients.
While significant work remains to scale the approach and close the gap to
optimisation oracles, the results establish a promising proof of
concept for AI-assisted environmental planning in critical
ecosystems.

\bibliographystyle{IEEEtran}
\bibliography{reference}

\appendices

\section{Scenario Ablations}\label{app:ablations}
We report two reward-design ablations alongside the headline
configuration of the main body. Unless noted, all three runs share the
same MaskablePPO algorithm, action masks (including the riparian
protection rule), and network architecture; they differ only in the
reward signal.

\begin{itemize}
    \item \textbf{Exp I - Eco-only}: $V_{\text{spatial}}$ zeroed. The reward
          reduces to $V_{\text{eco}}$ alone. For this ablation we use
          the SB3 PPO defaults rather than the tuned hyperparameters of
          the main body, since the single-term objective does not
          require the anneals or the $w_W$ curriculum.
    \item \textbf{Exp II - Spatial-without-regen}: full spatial reward
          (Eq.~\ref{eq:spatial}) with identical weights to the main
          body, but crop ESV held at its un-uplifted base of
          $\$246/\text{ha}/\text{yr}$ (normalised 0.22). This isolates
          the effect of the regenerative-agriculture coefficient
          uplift.
    \item \textbf{Exp III - Headline (Spatial+Regen)}: the main-body
          configuration, reproduced here for direct visual comparison.
\end{itemize}

\begin{figure}[htbp]
\centerline{\includegraphics[width=0.48\textwidth]{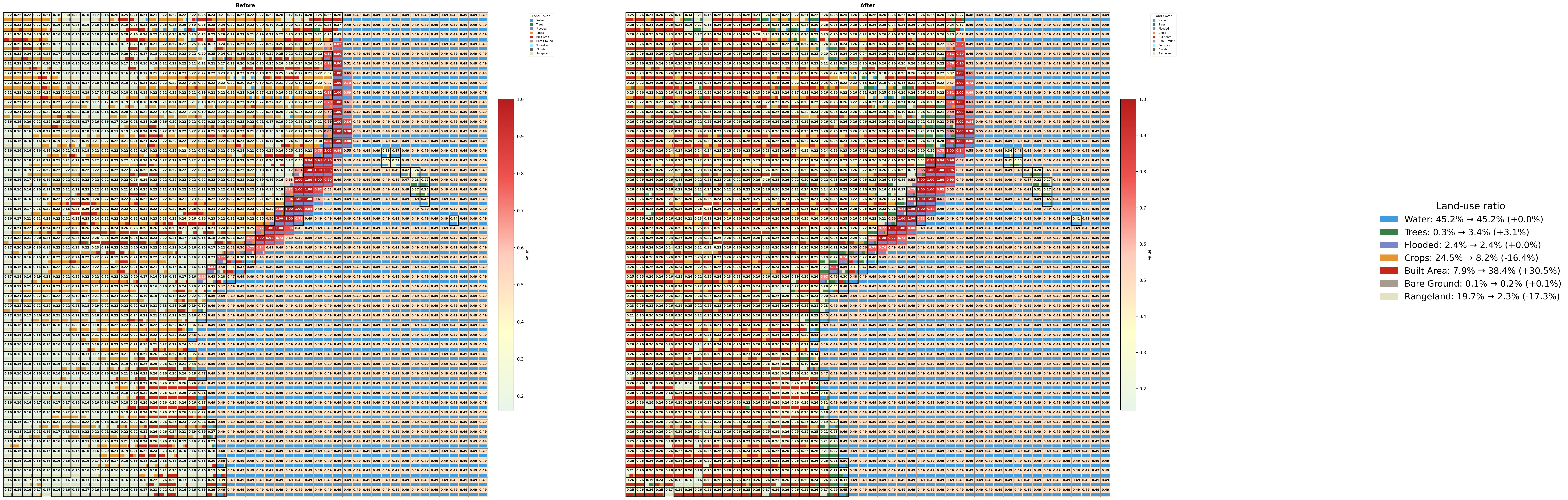}}
\caption{\emph{Eco-only} ablation. Without spatial shaping, the agent
         aggressively expands built area (the highest-value modifiable
         class), producing an ecologically unrealistic urban-sprawl
         allocation despite the riparian action mask still being
         active.}
\label{fig:abl_eco}
\end{figure}

\begin{figure}[htbp]
\centerline{\includegraphics[width=0.48\textwidth]{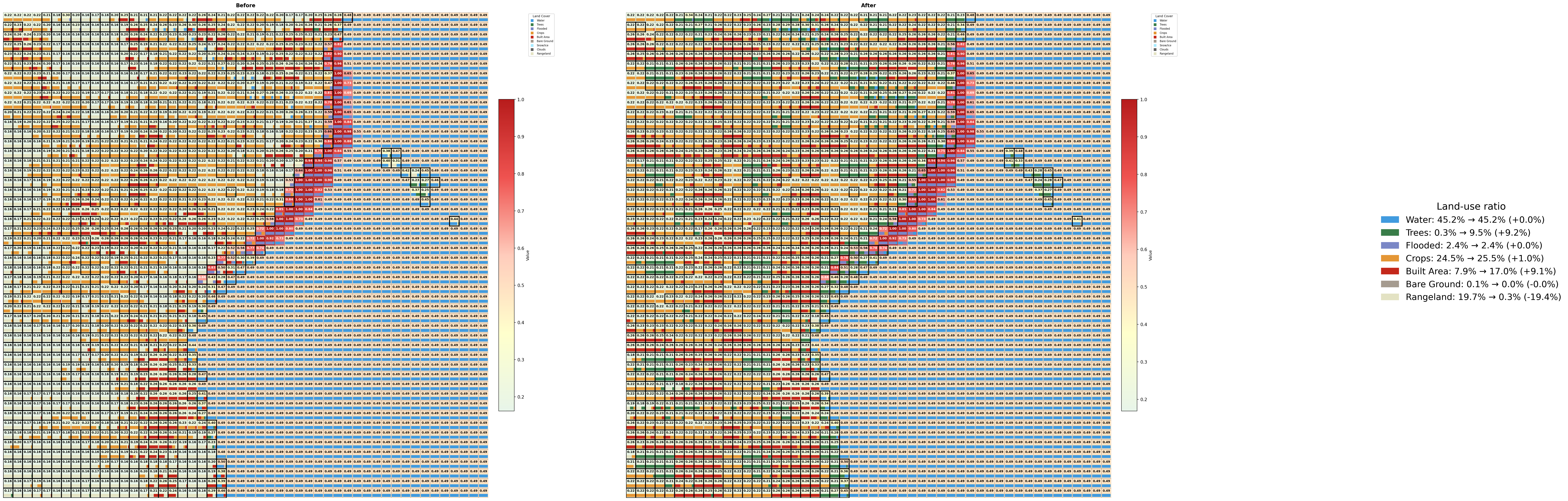}}
\caption{\emph{Spatial-without-regen} ablation. Spatial shaping yields
         forest consolidation, clustered built-area growth, and a
         restored riparian buffer; crop allocation remains moderate
         because base-crop ESV sits below built-area ESV.}
\label{fig:abl_spatial}
\end{figure}

Comparing Figs.~\ref{fig:abl_eco}--\ref{fig:abl_spatial} against the
headline Fig.~\ref{fig:primary} illustrates how the reward-design
components act: spatial shaping prevents urban sprawl and installs a
riparian buffer, while the regenerative-agriculture uplift shifts the
dominant conversion target from built area to crops.
Table~\ref{tab:comparison} summarises the qualitative behavioural
differences.

\begin{table}[htbp]
\caption{Qualitative Behaviour Across Reward-Design Ablations}
\label{tab:comparison}
\centering
\begin{tabular}{@{}p{1.5cm}p{1.7cm}p{1.8cm}p{1.8cm}@{}}
\toprule
 & \textbf{ExpI} & \textbf{ExpII} & \textbf{ExpIII} \\
 & \textbf{Eco-only} & \textbf{Spatial-without-regen} & \textbf{Headline (Spatial+Regen)} \\
\midrule
Dominant conversion &
  $\rightarrow$ Built &
  Mixed &
  $\rightarrow$ Crops \\[2pt]
Forest near water &
  Not obvious &
  Increased &
  Increased \\[2pt]
Built expansion &
  Aggressive &
  Clustered &
  Clustered, restrained \\[2pt]
Crop allocation &
  Moderate &
  Moderate &
  High \\[2pt]
Ecological realism &
  Low &
  High &
  High \\
\bottomrule
\end{tabular}
\end{table}

\section{Training Details and Loss-Curve Diagnostics}\label{app:training}
Table~\ref{tab:hyperparams} lists the full PPO hyperparameters used for
the spatially-aware runs (headline and spatial-without-regen); the
eco-only ablation uses SB3 defaults (no anneals, no curriculum).

\begin{table}[htbp]
\caption{PPO Hyperparameters (spatially-aware runs).}
\label{tab:hyperparams}
\centering
\begin{tabular}{@{}lc@{}}
\toprule
Parameter & Value \\
\midrule
Learning rate         & $5{\times}10^{-5} \rightarrow 5{\times}10^{-6}$ (last 33\%) \\
Rollout length ($n_{\text{steps}}$) & 2{,}048 \\
Mini-batch size       & 128 \\
PPO epochs            & 10 \\
Discount ($\gamma$)   & 0.99 \\
GAE $\lambda$         & 0.95 \\
Clip range            & 0.2 \\
Entropy coefficient   & $0.005 \rightarrow 0.001$ (last 33\%) \\
Value function coeff. & 0.5 \\
Max gradient norm     & 0.25 \\
Total timesteps       & 1{,}500{,}000 \\
Min initial total value (filter) & 1.0 \\
\bottomrule
\end{tabular}
\end{table}

\begin{figure}[htbp]
\centerline{\includegraphics[width=0.48\textwidth]{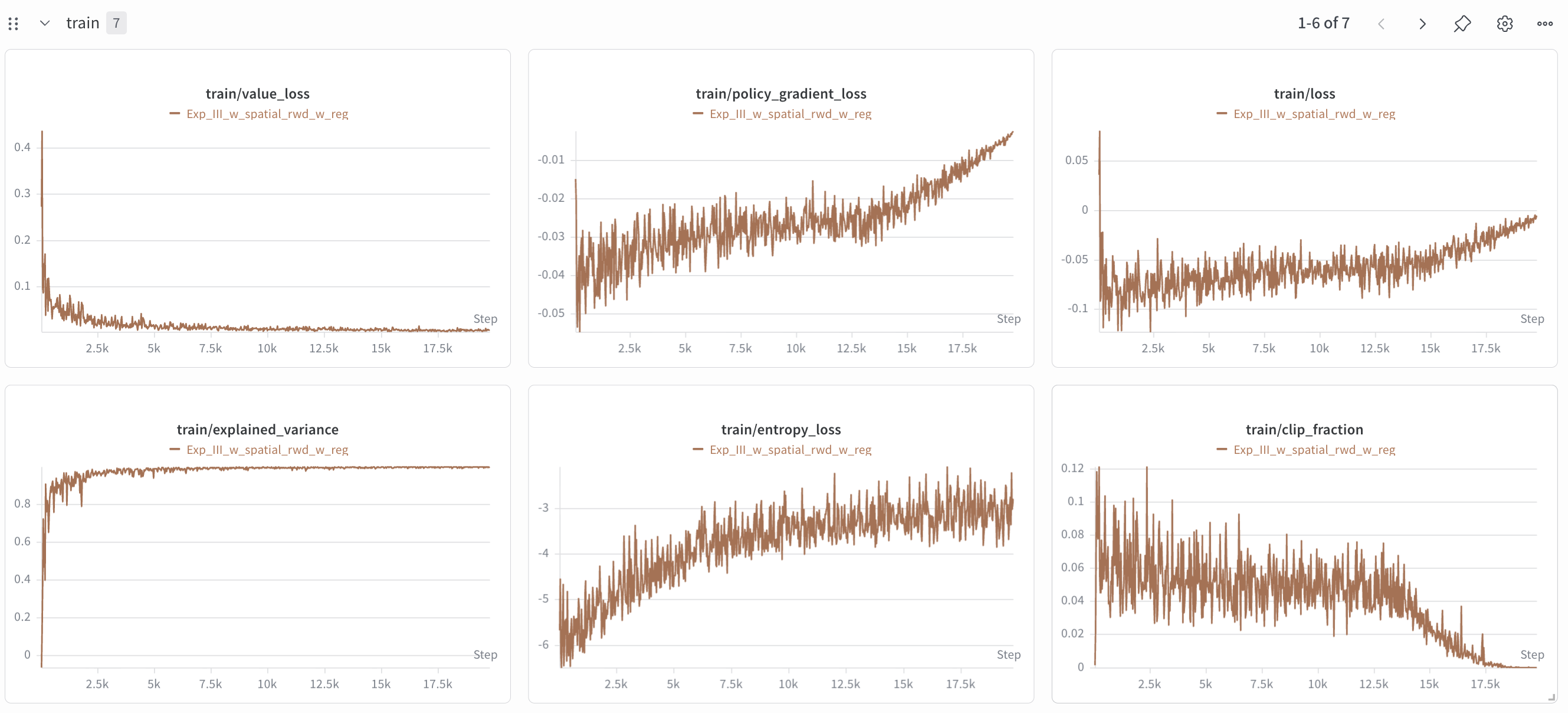}}
\caption{Training losses and diagnostics.}
\label{fig:metric_loss}
\end{figure}

\textbf{Critic fit and loss curves.}\quad
The critic converges rapidly:
\texttt{train/explained\_variance} saturates near $1.0$ within a few
thousand steps, while \texttt{train/value\_loss} drops sharply from
$\sim\!0.4$ to a small floor (Fig.~\ref{fig:metric_loss}). Mild
residual noise during the first $60\%$ of training tracks the $w_W$
curriculum, which non-stationarily shifts the value targets as the
buffer-penalty weight ramps from $1$ to $6$; the variance tightens
once the curriculum holds at $w_W{=}6$. The
policy-gradient loss drifts slightly more negative---
under-convergence rather than divergence, with budget left on the
table. \texttt{train/entropy\_loss} moves from roughly $-4.5$ toward
$-2$ nats (mean entropy $\sim\!4.5\!\to\!\sim\!2$, against the
theoretical maximum $\log 2500 \approx 7.82$~nats); the sharpening
in the final third reflects the entropy-coefficient anneal
($0.005\!\rightarrow\!0.001$). \texttt{train/clip\_fraction}
declines correspondingly from $\sim\!0.10$ to $\sim\!0.04$ as the
learning-rate decay ($5\!\times\!10^{-5}\!\rightarrow\!
5\!\times\!10^{-6}$) shrinks per-update step size. The trained
policy settles at $\sim\!25\%$ of maximum entropy---sharp enough
that deterministic and stochastic inference yield comparable rewards.

\textbf{Numerical handling of the 2500-dim softmax.}\quad
With the 2500-dim joint action space, fp32 accumulation error on the
softmax simplex ($\sim\!2500\times$ machine epsilon $\approx
3\!\times\!10^{-4}$) can exceed \texttt{torch}'s default $10^{-6}$
validation tolerance on \texttt{Categorical} construction. We disable
the strict check at the framework level; sampling and
\texttt{log\_prob} remain numerically sound, but a more principled fix
(log-space sampling or fp64 accumulation) would remove the need for
this workaround. This is a known issue with large \texttt{Discrete}
spaces in \texttt{torch}~2.x.

\section{Zoom-In Comparison Across Scenarios}\label{app:zoom}
\begin{figure*}[htbp]
\centerline{\includegraphics[width=0.95\textwidth]{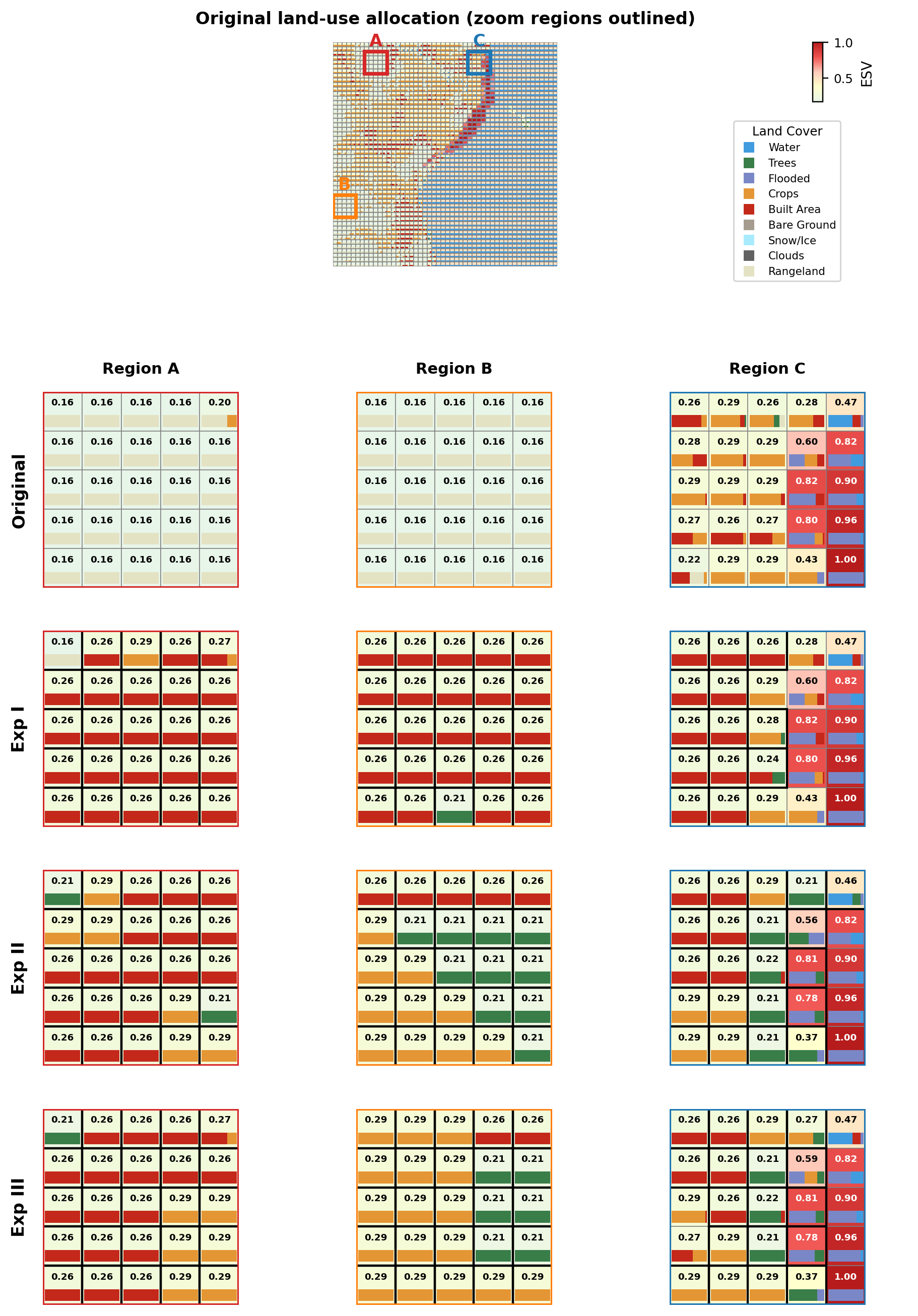}}
\caption{Zoom-in comparison across scenarios. \textbf{Top:} original
         land-use allocation with three coloured boxes marking the zoom
         regions. \textbf{Bottom:} per-region zooms for the original
         allocation and the final allocation under each reward-design
         scenario.}
\label{fig:zoom}
\end{figure*}

To make the per-cell allocation drift across the three scenarios
legible, Fig.~\ref{fig:zoom} pairs an overview of the original
allocation with a $4{\times}3$ zoom grid: rows correspond to the
original state and each of the three scenarios, while columns
correspond to three $5{\times}5$-cell example regions (outlined on the
overview in matching colours). All panels share a single ESV colormap
so cell-level value shifts are directly comparable. The black borders
inside each panel mark cells modified by the agent under that scenario.

\end{document}